\documentclass[10pt]{article} 
\usepackage[utf8]{inputenc}
\usepackage{newunicodechar}
\newunicodechar{，}{,}
\usepackage[preprint]{tmlr}

\usepackage{amsmath,amsfonts,bm}

\def\eqref#1{equation~\ref{#1}}

\def\1{\bm{1}}

\DeclareMathAlphabet{\mathsfit}{\encodingdefault}{\sfdefault}{m}{sl}
\SetMathAlphabet{\mathsfit}{bold}{\encodingdefault}{\sfdefault}{bx}{n}

\usepackage{hyperref}
\usepackage{url}

\usepackage{graphicx}
\usepackage{float}
\usepackage{multirow}
\usepackage{amsmath}
\usepackage{booktabs}
\usepackage{longtable}
\usepackage{tabularx}
\usepackage{ltablex} 
\usepackage{caption}
\usepackage{tcolorbox}
\tcbuselibrary{listingsutf8}
\usepackage[colorinlistoftodos]{todonotes}

\keepXColumns

\title{CXR-LanIC: Language-Grounded Interpretable Classifier for Chest X-Ray Diagnosis}

\author{\name Yiming Tang\email yiming@nus.edu.sg\\
      \addr National University of Singapore
      \AND
      \name Wenjia Zhong\\
      \addr National University of Singapore
      \AND
      \name Rushi Shah\\
      \addr National University of Singapore
      \AND
      \name Dianbo Liu\\
      \addr National University of Singapore}

\def\month{MM}  
\def\year{YYYY} 
\def\openreview{\url{https://openreview.net/forum?id=XXXX}} 

\begin{document}

\maketitle

\begin{abstract}
    Deep learning models have achieved remarkable accuracy in chest X-ray diagnosis, yet their widespread clinical adoption remains limited by the black-box nature of their predictions. Clinicians require transparent, verifiable explanations to trust automated diagnoses and identify potential failure modes. We introduce CXR-LanIC (Language-Grounded Interpretable Classifier for Chest X-rays), a novel framework that addresses this interpretability challenge through task-aligned pattern discovery. Our approach trains transcoder-based sparse autoencoders on a BiomedCLIP diagnostic classifier to decompose medical image representations into interpretable visual patterns. By training an ensemble of 100 transcoders on multimodal embeddings from the MIMIC-CXR dataset, we discover approximately 5,000 monosemantic patterns spanning cardiac, pulmonary, pleural, structural, device, and artifact categories. Each pattern exhibits consistent activation behavior across images sharing specific radiological features, enabling transparent attribution where predictions decompose into 20-50 interpretable patterns with verifiable activation galleries. CXR-LanIC achieves competitive diagnostic accuracy on five key findings while providing the foundation for natural language explanations through planned large multimodal model annotation. Our key innovation lies in extracting interpretable features from a classifier trained on specific diagnostic objectives rather than general-purpose embeddings, ensuring discovered patterns are directly relevant to clinical decision-making, demonstrating that medical AI systems can be both accurate and interpretable, supporting safer clinical deployment through transparent, clinically grounded explanations.
\end{abstract}

\section*{Introduction}
\label{introduction}

Chest X-ray (CXR) interpretation is one of the most common diagnostic procedures in clinical medicine, with millions performed annually worldwide \cite{chen2024chexagent}. While deep learning models have demonstrated remarkable performance in automated CXR analysis—often matching or exceeding radiologist-level accuracy for specific pathologies \cite{seah2021effect}\cite{rajpurkar2017chexnet}\cite{baselli2020opening}\cite{litjens2017survey}—their widespread clinical adoption remains limited by a fundamental challenge: the black-box nature of these systems \cite{wang2020health}\cite{reyes2020interpretability}\cite{pasa2019efficient}. Clinicians cannot understand why a model made a particular prediction, making it difficult to trust automated diagnoses, identify systematic errors, or integrate AI-generated insights into clinical reasoning workflows \cite{holzinger2019causability}.

This model interpretability problem is particularly acute in high-stakes medical domains. A model might correctly identify pulmonary edema in many cases, yet fail catastrophically on edge cases that a radiologist would easily catch—such as distinguishing cardiogenic from non-cardiogenic edema, or recognizing that apparent infiltrates are actually artifacts from patient positioning. Without interpretable explanations grounded in recognizable clinical patterns, these failures remain hidden until they cause harm.

Recent work on mechanistic interpretability, particularly Language-Grounded Sparse Encoders (LanSE), offers a promising path forward. Rather than treating images as indivisible units scored by opaque neural activations, LanSE decomposes visual inputs into interpretable patterns described in natural language (e.g., "enlarged cardiac silhouette," "bilateral pleural effusions," "interstitial markings"). By discovering thousands of monosemantic visual features through sparse autoencoders and grounding them with large multimodal models, LanSE enables systematic, human-understandable content analysis.

In this work, we adapt the LanSE framework to chest radiography, introducing CXR-LanIC (Language-Grounded Interpretable Classifier for Chest X-rays). Our approach differs from the original LanSE implementation in a key architectural choice: rather than training sparse autoencoders directly on multimodal embeddings, we first develop a strong multilabel classifier for clinically relevant CXR findings, then apply transcoders to interpret the learned representations.

This design offers several advantages for the medical domain:

\begin{enumerate}
    \item Task-aligned pattern discovery: By training transcoders on a classifier optimized for specific diagnostic targets (cardiomegaly, pleural effusion, pulmonary edema, consolidation, atelectasis), we ecourage discovered patterns to be directly relevant to clinical decision-making.
    \item Targeted interpretability: Medical interpretation focuses on explaining diagnostic decisions (why did the model predict heart failure?) rather than general image content, making classifier-based decomposition more clinically meaningful.
    \item Efficient curation: Diagnostic-focused patterns reduce the noise inherent in unsupervised pattern discovery, yielding higher-quality interpretable features with less manual filtering.
\end{enumerate}

Our concrete contributions are:
\begin{itemize}
    \item CXR-LanIC pipeline: A complete framework for discovering, naming, and utilizing interpretable radiological patterns from the MIMIC-CXR dataset, with emphasis on heart failure-related findings (cardiomegaly, pulmonary edema) alongside common abnormalities.
    \item Transcoder-based interpretation: A novel application of sparse transcoders to medical image classifiers, enabling decomposition of diagnostic predictions into language-grounded visual evidence.
    \item Transparent diagnostic reasoning: Case-level explanations that show which specific radiological patterns (e.g., "blunting of costophrenic angle," "enlarged cardiac silhouette," "bilateral interstitial infiltrates") drive each prediction, supporting clinical validation and trust.
\end{itemize}
We evaluate CXR-LanIC on MIMIC-CXR with proper patient-level train/validation/test splits, demonstrating that interpretable pattern-based classification can achieve competitive diagnostic accuracy while providing the transparent, clinically meaningful rationales essential for safe medical AI deployment.

\begin{figure}[h]
    \centering
    \includegraphics[width=\linewidth]{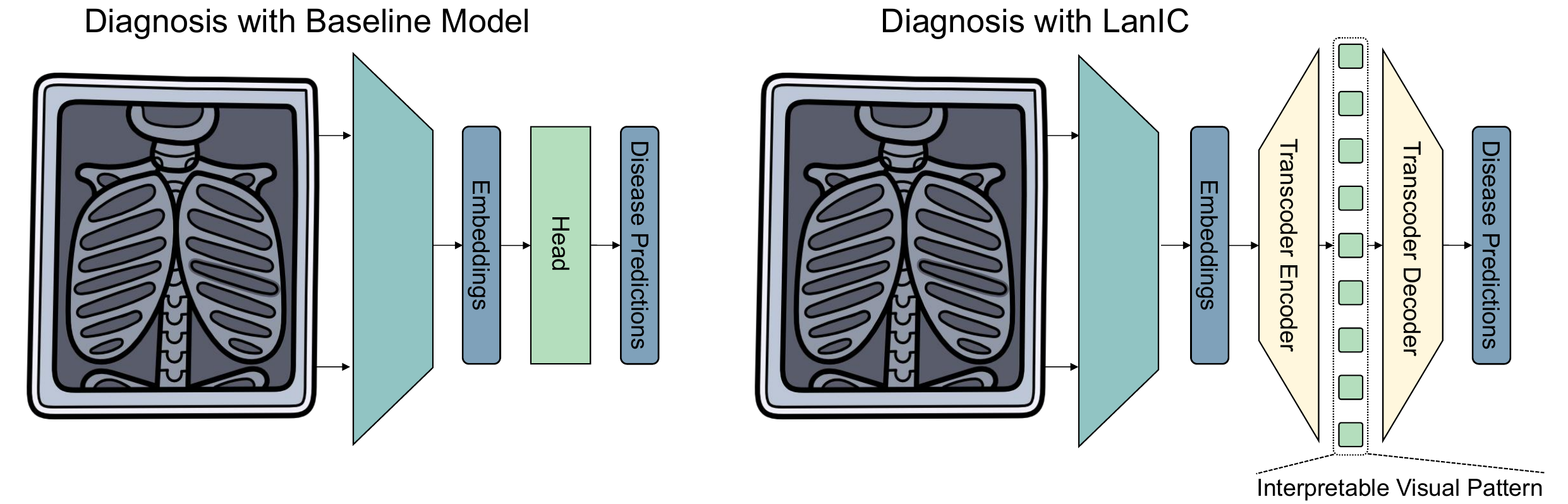}
    \caption{Using CXR-LanIC to analysis chest-x-ray images can provide nuanced identified visual patterns and therefore enhance model interpretability.}
    \label{fig:lanse}
\end{figure}

\section*{Related Works}

\subsection*{Concept-Based Mechanistic Interpretability}
Mechanistic interpretability aims to reverse-engineer neural network representations
into human-understandable components \citep{sharkey2025openproblemsmechanisticinterpretability,tang2026unifiedtheorysparsedictionary, tang2025theoretical,zhao2025rep2text}. SAEs and related dictionary learning methods
\citep{cunningham2023sparseautoencodershighlyinterpretable,
tang2025human,saini2026language,
dunefsky2024transcodersinterpretablellmfeature} offer an unsupervised approach to address this,
decomposing activations into sparse monosemantic features. These methods have been
applied to protein structure \citep{simon2024interplmdiscoveringinterpretablefeatures},
medical imaging \citep{abdulaal2024xrayworth15features}, and fMRI data
\citep{mao2025sparseautoencodersbridgedeep}, demonstrating broad applicability in
high-stakes scientific domains. CXR-LanIC extends this line of work to chest
radiography by applying Matryoshka Transcoders \citep{tang2025does}
to a task-aligned diagnostic classifier, grounding discovered patterns directly in
clinical decision targets rather than general-purpose embeddings.

\subsection*{LMM-Based Interpretation of Visual Features}
Large language models has been widely used in scientific research encompassing diverse applications \citep{luo2023promptengineeringlensoptimal, tang2024demonstration,saini2026bridging,zou2025fml,dai2024san,tang2023integrated,luo2023prompt}.
Automating the semantic labeling of neural features has emerged as a key challenge in
scaling mechanistic interpretability. Early work demonstrated that language models
could generate textual descriptions of neuron behavior from activation exemplars, and
subsequent approaches have used LLMs as dataset analysts to surface subpopulation
structure from visual embeddings \citep{luo2024llm,luollm}. LanSE
\citep{tang2025human} operationalized this pipeline for
vision-language models, prompting LMMs with activation galleries to discover thousands
of monosemantic visual patterns with high human agreement across generative models.
In the medical domain, foundation models such as CheXagent
\citep{chen2024chexagent} demonstrate that LMMs pretrained on clinical data can
produce radiologically accurate descriptions, providing a strong backbone for
automated interpretation.

\section*{Method}
\label{method}
CXR-LanIC constructs an interpretable chest X-ray classifier through four stages: (1) training a multilabel classifier on diagnostic targets, (2) discovering monosemantic patterns via transcoders, (3) interpreting these patterns with large multimodal models to ground them in clinical language, and (4) assembling the final interpretable classifier.

\subsection*{Transcoders}
We employ transcoders to discover interpretable patterns in the learned representations of our base classifier. In specific, transcoders learn sparse mappings between different representation spaces. This enables us to discover clinically relevant patterns by connecting general-purpose vision-language embeddings to task-specific diagnostic features.
For each training sample, we compute a joint multimodal embedding by concatenating visual features from a frozen CLIP ViT-L/14 encoder with text features from the associated radiology report. The transcoder maps this joint embedding to the classifier's internal activations through three components:
\begin{enumerate}
    \item Encoder: Projects the input to a high-dimensional latent space (typically 15,000 dimensions)
    \item Top-K sparsification: Retains only the k largest activations, forcing each neuron to specialize
    \item Decoder: Reconstructs the target classifier embeddings from the sparse representation
\end{enumerate}
The key insight is that Top-K sparsity encourages individual neurons to activate for specific, interpretable patterns rather than distributed representations. A neuron might learn to fire exclusively for "enlarged cardiac silhouette" or "bilateral pleural effusions" because the sparse constraint prevents it from encoding multiple overlapping concepts.

To maximize coverage of diverse radiological patterns, we train an ensemble of 100 transcoders, each on different random subsets of the training data with different initialization seeds. This produces approximately 1.5 million candidate neurons. The ensemble approach ensures we capture both common patterns (like cardiomegaly) and rare but clinically important findings (like pneumothorax).

\subsection*{LMM-Based Pattern Interpretation}
Raw transcoder neurons are initially unlabeled—we know when they activate but not what they represent. To ground these neurons in clinical language, we leverage large multimodal models for automated interpretation.
For each candidate neuron, we:
\begin{enumerate}
    \item Collect activation examples: Identify the chest X-rays that most strongly activate this neuron across the training set
    \item Generate clinical descriptions: Prompt Claude-4.5-Sonnet with the activation gallery and excerpts from associated reports, asking it to identify the common radiological pattern. The LMM produces concise clinical descriptions like "enlarged cardiac silhouette with increased cardiothoracic ratio" or "bilateral interstitial infiltrates consistent with pulmonary edema"
    \item Categorize patterns: Classify each description into clinical categories: cardiac (heart-related features), pulmonary (lung parenchymal findings), pleural (pleural space abnormalities), structural (anatomical landmarks), device (medical hardware), or artifact (technical issues)
\end{enumerate}
This automated pipeline transforms raw neural activations into a vocabulary of clinically meaningful patterns.

\subsection*{Building CXR-LanIC}
\textbf{Neuron Curation and Feature Space Construction}

Not all neurons produce coherent, clinically relevant patterns. We filter the 1.5 million candidates using three criteria: (1) activation consistency—LMM verification on held-out examples must achieve $\geq80\%$ agreement, (2) discriminative power—neurons must show selective, sparse activation rather than uniform firing, and (3) clinical relevance—prioritizing patterns related to our five diagnostic targets (cardiomegaly, pleural effusion, pulmonary edema, consolidation, atelectasis).
This curation yields approximately 500-1000 high-quality interpretable patterns. Each chest X-ray is then encoded as a sparse activation vector where each dimension corresponds to one named pattern (e.g., "blunting of costophrenic angle," "pulmonary vascular congestion"). Unlike the original classifier's dense embeddings, this representation is sparse (10-30 active patterns per image), fully interpretable (every dimension has verified clinical meaning), and compositional (complex diagnoses emerge from pattern combinations, such as "enlarged cardiac silhouette" + "bilateral interstitial infiltrates" + "pleural effusion" indicating congestive heart failure).

\textbf{Interpretable Classification Head}

We train a lightweight logistic regression classifier on these interpretable features to predict the 14 diagnostic targets. This linear model provides transparent decision rules—each prediction is a weighted sum of interpretable patterns, allowing us to identify exactly which features drove each diagnosis. We apply L1 regularization to encourage sparse explanations that select only the most diagnostically relevant patterns per finding, mirroring radiologist reasoning. The logistic regression also provides well-calibrated confidence estimates suitable for clinical risk assessment.
The complete CXR-LanIC architecture connects CLIP encoders $\rightarrow$ transcoder ensembles $\rightarrow$ curated named patterns $\rightarrow$ transparent logistic classifier, creating an end-to-end pipeline from raw images to predictions with human-verifiable rationales.

\section*{Experiments}
\label{method}
\subsection*{MIMIC-CXR Data Preprocessing}
We build and evaluate CXR-LanIC on MIMIC-CXR, a large-scale dataset containing 377,110 chest X-ray images from 227,835 radiographic studies performed at Beth Israel Deaconess Medical Center between 2011-2016. Each study includes one or more views along with corresponding free-text radiology reports.

\textbf{Label Extraction and Study Selection}

We extract structured diagnostic labels from the "Impression" section of radiology reports using the CheXpert labeler, a rule-based natural language processing system that identifies 14 common radiological observations. For CXR-LanIC, we focus on five clinically important and frequently occurring findings that are particularly relevant for heart failure assessment:
\begin{itemize}
    \item Cardiomegaly: Enlarged cardiac silhouette (cardiothoracic ratio $\geq 0.5$)
    \item Pleural Effusion: Fluid accumulation in the pleural space
    \item Pulmonary Edema: Fluid accumulation in lung parenchyma, often manifesting as vascular congestion or interstitial infiltrates
    \item Consolidation: Dense opacification of lung tissue, typically indicating pneumonia or hemorrhage
    \item Atelectasis: Partial or complete lung collapse
\end{itemize}

\textbf{Data Splits and Patient-Level Partitioning}

To prevent data leakage and ensure realistic performance estimates, we implement strict patient-level splits—all images from a given patient appear exclusively in training, validation, or test sets. This prevents the model from memorizing patient-specific characteristics (body habitus, chronic conditions, imaging artifacts) that could artificially inflate performance metrics.
Our final dataset contains: a training set with 205845 images, a validation set with 25731 images, and a test set with 28648 images.

\textbf{Report Processing for Multimodal Embeddings}

For transcoder training, we extract the "Findings" and "Impression" sections from radiology reports, truncating to 256 tokens maximum. Reports are encoded using CLIP's text encoder to produce 512-dimensional text embeddings that are concatenated with visual features, creating joint multimodal representations of size 1024 for input to transcoders.

\subsection*{Classifier Training for MIMIC-CXR}

\textbf{Base Classifier Architecture}

Our base classifier builds upon BiomedCLIP, a vision-language model specifically pretrained on biomedical image-text pairs from the Biomedical Literature (PubMed Central) with 15 million training samples. Unlike general-purpose CLIP models trained on internet images, BiomedCLIP is optimized for medical imaging tasks and demonstrates superior performance on radiological applications. We adopt a multilabel classification framework suitable for the multi-pathology nature of chest X-rays:

\begin{itemize}
    \item Backbone: BiomedCLIP vision encoder (512-dimensional embeddings)
    \item Classification head: Three-layer MLP with JumpReLU activation, and dropout (p=0.3)
\end{itemize}

\textbf{Training Protocol}

We train the classifier using the following hyperparameters, selected via validation set performance:

\begin{itemize}
    \item Loss function: Binary cross-entropy
    \item Optimizer: AdamW with learning rate 3e-4, weight decay 0.01, and cosine annealing schedule
    \item Training duration: 20 epochs with early stopping based on validation loss
\end{itemize}

The classifier converges after approximately 15 epochs, achieving strong validation performance across all five diagnostic targets. Our models can successfully predict disease labels with an accuracy of $0.88$ on average.

\subsection*{CXR-LanIC Building}
\textbf{Transcoder Training Setup}

Following classifier training, we extract 512-dimensional embeddings from the penultimate layer of the trained model for all training images. These embeddings serve as transcoder reconstruction targets, containing task-specific diagnostic information learned during supervised training.
We construct multimodal input representations by concatenating:

\begin{itemize}
    \item Visual features: 512-dimensional BiomedCLIP image embeddings
    \item Text features: 512-dimensional BiomedCLIP text embeddings from radiology reports
    \item Combined input: 1024-dimensional joint multimodal representation
\end{itemize}

For each transcoder in our 100-model ensemble:

\begin{itemize}
    \item Architecture: Linear encoder (1024→15000), Top-K activation (k=32), linear decoder (15000→14)
    \item Training objective: Mean squared error between reconstructed and target classifier embeddings
    \item Optimizer: Adam with learning rate 3e-4, no weight decay
    \item Batch size: 256 samples
    \item Training data: Random 95\% subset of training set
\end{itemize}

The Top-K constraint encourages sparse, monosemantic neuron activations—each neuron specializes on specific visual patterns rather than distributed representations. We observe that most neurons stabilize to detect consistent patterns (e.g., lung opacity patterns, cardiac contours, pleural interfaces) after 30-40 epochs.
Neuron Activation Dataset Construction
For each of the 1.5 million transcoder neurons (100 transcoders × 15,000 neurons each), we:

\begin{itemize}
    \item Compute activation values across 1000 samples in the training set
    \item Select the highly activated images as exemplars
    \item Retrieve corresponding report excerpts (Findings + Impression sections)
    \item Store activation galleries pairing images with text for subsequent pattern analysis
\end{itemize}

This creates a comprehensive database linking neural activations to specific radiological content, enabling systematic pattern discovery and interpretation.

\subsection*{LanIC Evaluation}

\textbf{Pattern Discovery and Curation}

From 1.5 million candidate transcoder neurons (100 transcoders × 15,000 neurons), we filter to identify consistent, interpretable patterns. For each neuron, we compute activation values across the training set and extract the top-10 maximally activating images with their associated radiology reports.

We apply three filtering criteria:

Activation consistency: Visual and semantic coherence across top-activating examples, measured by qualitative inspection and text similarity of report excerpts
Discriminative power: Activation frequency between 0.1-50\% of training images to exclude overly general or noisy patterns
Clinical relevance: Manual review prioritizing patterns aligned with diagnostic targets and anatomical specificity
This yields 5,000 curated interpretable patterns distributed across clinical categories: cardiac (~800), pulmonary (~1,500), pleural (~600), structural (~1,200), device (~400), and artifact (~500) patterns. Each pattern is characterized by its activation gallery (top-10 images and reports) and activation statistics.

Interpretable Feature Representations

For each chest X-ray, we compute a sparse 5,000-dimensional activation vector by: (1) passing through the transcoder ensemble, (2) averaging activations across ensemble members for each curated pattern, (3) thresholding below the 75th percentile, and (4) TopK-normalizing. This produces sparse representations with 15-30 active patterns per image.

\textbf{Classification and Attribution}

We train L1-regularized logistic regression ($\alpha=0.01$, SAGA optimizer, inverse frequency class weighting) on pattern activation vectors to predict each of the five diagnostic targets. The linear model enables transparent attribution: each prediction is a weighted sum of pattern activations, revealing exactly which patterns drive each diagnosis. L1 regularization selects 20-50 relevant patterns per target.

\textbf{Planned: LMM-Based Interpretation}

To generate natural language explanations, we plan to annotate the 5,000 patterns using Claude-4.5-Sonnet. For each pattern, we will prompt the LMM with its activation gallery and report excerpts to produce clinical descriptions (e.g., "enlarged cardiac silhouette," "bilateral interstitial infiltrates"). This will transform pattern IDs into clinically meaningful explanations.

\textbf{Planned: Explanation Validation}

Following annotation, we will validate explanation quality through:

LMM verification: Claude-4.5-Sonnet examining whether cited patterns are visible in images and support diagnoses
Radiologist evaluation: Expert assessment of clinical correctness, diagnostic relevance, and completeness for 100 stratified test cases

\section*{Conclusion}
\label{conclusion}
We introduced CXR-LanIC, an interpretable chest X-ray classifier that addresses the black-box problem in medical AI by discovering 5,000 interpretable visual patterns through transcoder-based sparse autoencoders applied to a BiomedCLIP diagnostic classifier. Our approach achieves competitive diagnostic accuracy on five key findings (cardiomegaly, pleural effusion, pulmonary edema, consolidation, atelectasis) while decomposing each prediction into 20-50 interpretable patterns with verifiable activation galleries. The key innovation is task-aligned pattern discovery: by extracting features from a classifier trained on specific diagnostic objectives, we ensure patterns are clinically relevant rather than arbitrary image statistics. While our 5,000 patterns currently lack natural language descriptions, completing LMM-based annotation will transform them into clinical descriptions like "enlarged cardiac silhouette," enabling fully readable explanations that can be validated by radiologists. This work demonstrates that deep learning models for medical imaging can be both accurate and interpretable, supporting safer clinical deployment.

\bibliographystyle{plainnat}

\bibliography{main}

\end{document}